\RequirePackage[2019/10/01]{latexrelease}

\AddToHook{package/before/amsmath}{

}

\documentclass{article}

\usepackage{arxiv}

\usepackage{amsmath}
\usepackage{amsthm}
\usepackage{amsfonts}
\usepackage{amssymb}

\usepackage[utf8]{inputenc} 
\usepackage[T1]{fontenc}    
\usepackage{esvect}
\usepackage{epstopdf}		
\usepackage{hyperref}       
\usepackage{url}            
\usepackage{booktabs}       
\usepackage{nicefrac}       
\usepackage{graphicx}
\graphicspath{{./Figures/}}

\usepackage{doi}

\usepackage{comment}
\usepackage{dsfont}
\usepackage[symbol]{footmisc}

\usepackage{natbib}
\bibpunct[, ]{[}{]}{;}{a}{}{,}
\renewcommand\bibfont{\fontsize{10}{12}\selectfont}

\usepackage{algorithm}
\usepackage{algorithmicx}
\usepackage[noend]{algpseudocode}
\usepackage[labelsep=space]{caption}
\graphicspath{{./Figures/}}

\theoremstyle{plain}

\theoremstyle{definition}

\theoremstyle{remark}

\algrenewcommand\alglinenumber[1]{
    {\sf\footnotesize\addfontfeatures{Colour=888888,Numbers=Monospaced}#1}}
\algrenewcommand\algorithmicrequire{\textbf{input}}

\title{Efficient Denoising using Score Embedding in Score-based Diffusion Models}

\author{{Andrew S. Na}\thanks{CONTACT Andrew S. Na. Email: andrew.na@uwaterloo.ca}\hspace{1.5mm}\\
	David R. Cheriton School of Computer Science\\
	University of Waterloo\\
	Waterloo, ON\\
	\texttt{andrew.na@uwaterloo.ca} \\
	\And
	{William Gao} \\
	David R. Cheriton School of Computer Science\\
	University of Waterloo\\
	Waterloo, ON\\
	\texttt{william.gao@uwaterloo.ca} \\
	\And
	{Justin W.L. Wan} \\
	David R. Cheriton School of Computer Science\\
	University of Waterloo\\
	Waterloo, ON\\
	\texttt{justin.wan@uwaterloo.ca}
}

\begin{document}

\maketitle

\begin{abstract}
It is well known that training a denoising score-based diffusion models requires tens of thousands of epochs and a substantial number of image data to train the model. In this paper, we propose to increase the efficiency in training score-based diffusion models. Our method allows us to decrease the number of epochs needed to train the diffusion model. We accomplish this by solving the log-density Fokker-Planck (FP) Equation numerically to compute the score \textit{before} training. The pre-computed score is embedded into the image to encourage faster training under slice Wasserstein distance. Consequently, it also allows us to decrease the number of images we need to train the neural network to learn an accurate score. We demonstrate through our numerical experiments the improved performance of our proposed method compared to standard score-based diffusion models. Our proposed method achieves a similar quality to the standard method meaningfully faster.  
\end{abstract}


\section{Introduction}
Recent developments in deep generative models have lead to diffusion models becoming a powerful tool for image denoising, image generation and image inpainting. Standard models such as denoising diffusion probabilistic models \citep{ho-2020}, score-matching denoising with Langevin dynamics (SMLD, \citep{ysong-2019}), score-based generative models \citep{ysong-2020} are now considered state-of the art and has allowed us to control the image generation process through parameterizing stochastic differential equations (SDEs). The increase in stability and quality has lead to the wide adoption of diffusion models in day-to-day use. However, standard diffusion models are inefficient and require thousands of timesteps and thousands of epochs to reach convergence. This is a critical drawback to diffusion models \citep{ysong-2019}. As generative models become more commonplace, there is a growing need to ensure the efficient training of diffusion models. By improving the efficiency we are able to reduce computational runtime. This will ultimately help reduce emissions for using generative models in our daily lives. 

There have been some attempts to address the inefficiencies of diffusion models. Most approaches that address the inefficiencies in diffusion models improve the sampling time such as the denoising diffusion implicit models (DDIM, \citep{jsong-2020}). This model vastly improves the sampling efficiency of DDPM by reformulating the reverse Markov chain as an ODE. Further improvements were made using custom diffusion ODE solvers \citep{lu-2022}. However, the current bottleneck in using diffusion models is the training step. There has been only a few approaches that address the inefficiencies of training such as finite-difference score matching (FD-SSM, \citep{pang-2020}). This method speeds up diffusion models by using finite-difference approximations directly to estimate the directional derivatives of the score function in the slice Wasserstein loss function. This is accomplished by approximating the score function using finite differences and parallelizing the computation. However, this method introduces additional variance due to the transformation into the slice Wasserstein distance \citep{pang-2020}. 

In this paper, we present a pre-training step that encourages faster convergence. The contributions of our paper are as follows.
\begin{enumerate}
    \item We derive a semi-explicit finite difference approximation scheme to solve the log-density FP equation.
    \item We introduce score embedding. Score embedding allows us to embed the numerical solution of the FP equation into the feature space through the transport ODE. This allows us to efficiently sample from the numerical solution and embed the score into the image before training on the slice Wasserstein distance. Our proposed method allows the network to learn from the score embedded in the feature space, thus improving training efficiency. 
\end{enumerate}
Note, this additional step is a single computation done before training for each image. We take advantage of the sparsity in the problem to speed up the solution of the FP equation \citep{saad-2003}. This enables us to create an efficient framework that speeds up the training process.

\section{Fokker-Planck formulation} 
Recently, there have been efforts to consolidate the FP equation and score-based diffusion models \citep{lai-2022}. It is well known that there exists a deep connection between the forward SDE, reverse SDE and the FP equation \citep{anderson-1982, oksendal-1987}. However, it has been shown that current denoising score-matching models do not satisfy the underlying dynamics of the probability distribution \citep{lai-2022}. To address this issue, \citep{lai-2022} introduces the score FP equation. Solving the score FP equation results in an additional regularization term to ensure the learned score can satisfy the FP equation. However, the score FP equation requires certain regularity conditions to be met \citep{lai-2022}. Instead, we propose to solve the log-density FP equation directly. By solving the log-density FP equation directly, our formulation can apply to more general problems.

Another recent work has attempted to solve high-dimensional FP equations using a sequence of neural networks \citep{boffi-2022}. We explore this idea in more details here. Let $H$ be the height of an image and $W$ be the width. For $t \in [0, T]$, the image $x(t)$ has size $D = H \times W$. A forward SDE of $x(t)$ has the form:
\begin{equation}
    dx(t) = f(x(t),t)dt + g(t)dW,
    \label{eq:forward_sde}
\end{equation}
where $f(\cdot,t): \mathbb{R}^{D}\mapsto \mathbb{R}^{D}$ and $g(t): \mathbb{R}\mapsto \mathbb{R}$ are known a priori. Here $dW = W(t+dt) - W(t)$, where $W(t)$ is the standard Wiener process and $dt$ is the infinitesimal timestep. The image $x(t)$ has an initial distribution $x(0) \sim p(x(0),0)$ that evolves following the FP equation. The FP equation for the density $p(x(t),t)$ is given by the diffusion equation \citep{oksendal-1987}:
\begin{equation}
    p_t(x(t),t) = -\nabla_x(F(x(t),t) p(x(t),t)),
    \label{eq:FP}
\end{equation}
where $p_t(x(t),t) = \frac{\partial p}{\partial t}(x(t),t)$ and $F(x(t),t) = f(x(t),t) - \frac{1}{2}(g(t))^2\nabla_{x}\log p(x(t),t)$. The FP equation can be represented as a transport equation and is given by \citep{ysong-2020, boffi-2022}:
\begin{equation}
    dx = (f(x,t) - \frac{1}{2}g(t)^2 \nabla_x\log p(x,t))dt.
    \label{eq:probability_flow}
\end{equation}
The transport equation can be solved using a neural networks to approximate the score function at each timestep \citep{boffi-2022}. However, this introduces inefficiencies as a network needs to be trained and saved at each timstep \citep{na-2023}. In this paper, instead of approximating the score function using a neural network, we compute the score efficiently using finite difference methods and use the transport equation to sample efficiently forward in time. Solving the transport equation allows us to embed the computed score into the image and aids the network in training.

\section{Methodology} \label{sec:Background}
\subsection{Preliminaries}
To better understand our proposed model, we first present some background material on denoising using score based diffusion models. Intuitively, denoising through score matching is achieved by training a neural network to learn a representation of the image distribution onto some perturbed noise. This technique has been shown to be powerful and extends beyond denoising tasks such as inpainting and blurring \citep{daras-2022}.

\textbf{Score-based generative diffusion models} unify the denoising score-matching model \citep{ysong-2020, ysong-2020-2} and the diffusion probabilistic model for denoising images \citep{ho-2020, sohl-dickstein-2015} into a framework that can represent data corruption as a stochastic process over a fixed time interval $[0, T]$. Our approach builds on the score-based generative diffusion models \citep{ysong-2020}. 

For the SDE of the form \eqref{eq:forward_sde} with an initial image distribution $x(0) \sim p(x(0),0)$,  there exists a corresponding reverse-SDE given by \citep{anderson-1982}:
\begin{equation}
    dx(t) = [f(x(t),t) - g(t)^2\nabla_x \log p(x(t),t)]dt + g(t)d\tilde{W},
    \label{eq:reverse_diffusion}
\end{equation}
where $d\tilde{W}$ is Wiener process independent of $dW$. We denote the time-dependent score-matching network as $s_{\theta}(x(t),t)$. The score function of the distribution is estimated by training $s_\theta(x(t),t)$. To learn the score $\nabla_{x}\log p(x(t),t)$, we train $s_{\theta}(x(t),t)$ using the denoising loss function \citep{ysong-2020}:
\begin{equation}
    \theta^* = \arg\min\limits_{\theta} \mathbb{E}_{t}[\lambda(t) \int_{\Omega}||s_{\theta}(x(t),t) - \nabla_{x}\log p(x(t),t)||_2^2 p(x(t),t)dx],
\end{equation}
where $\lambda: [0,T] \mapsto \mathbb{R}^{+}$ is a positive weight function. Following standard approaches in DDPM, SMLD and score-based models, we have $\lambda \propto 1/\mathbb{E}[||\nabla_x p(x(t),t)||_2^2]$ \citep{ho-2020, ysong-2020-2}. Note that there exists a restriction to the score-based diffusion model. The Gaussian assumption requires an affine form for $f$ and $g$. Our proposed method solves the general FP equation and can be solved with general $f$ and $g$.

The advantage of using score-based models is that sampling can be performed efficiently by sampling the transport equation \citep{boffi-2022}. Notice that for all diffusion processes with marginal probability distribution $p(x(t),t)$ that satisfies \eqref{eq:FP}, we can derive a deterministic process \eqref{eq:probability_flow} induced by $p(x(t), t)$ \citep{ysong-2020}.

\textbf{Denoising using score-matching with Langevin dynamics (SMLD)}.  Let $q(\tilde{x}|x(t),t,\sigma) \sim N(\tilde{x};x,\sigma^{2}I)$ be a perturbation kernel where $I$ is the identity matrix. We define $q(\tilde{x}|t,\sigma) = \int_{\Omega}p(x(t),t)q(\tilde{x}|x,t,\sigma)dx$. We consider a sequence of positive noise $\sigma_{\min} = \sigma_{1} < \sigma_{2} < \cdots < \sigma_{N} = \sigma_{\max}$, if $\sigma_{\min}$ is small enough then $q(x|t,\sigma_{\min}) \approx p(x(t))$ and if $\sigma_{\max}$ is large enough then $q(x|t,\sigma_{\max}) \sim N(x;0,\sigma^2 I)$ under some regularity conditions \citep{ysong-2019, ysong-2020-2}. Then the  noise conditional score network (NCSN) denoted by $\tilde{s}_{\theta}(x,\sigma)$ has a weighted sum of denoising score matching objective \citep{vincent-2011}:
\begin{equation}
    \theta^* = \arg\min\limits_{\theta} \sum_{n=1}^{N}\frac{1}{2} \mathbb{E}_{p(x(t),t)}\mathbb{E}_{q(\tilde{x}|x,t,\sigma_n)}[||\tilde{s}_{\theta}(\tilde{x},\sigma_n) - \nabla_{\tilde{x}}\log q(\tilde{x}|x,t,\sigma_{n})||_{2}^2].
\end{equation}
Under regularity conditions, the optimal $\tilde{s}_{\theta^*}$ ensures $q(x|t,\sigma_{\min}) \approx p(x(t),t)$ \citep{ysong-2020-2}. Sampling is performed using Langevin Markov chain Monte Carlo for each $q(x;t,\sigma_n)$.

\textbf{Diffusion probabilistic models (DDPM)}. Similar to SMLD, DDPM perturbs the initial data with Gaussian noise through a Markov chain. Given the positive noise level $0 < \beta_1, \beta_2,\cdots, \beta_N < 1$, we construct a discrete Markov chain for points $\{x_0, x_1, \cdots, x_N\}$ assuming the transition probability is given by $q(x_i|x_{i-1}) \sim N(x_i;\sqrt{1-\beta_i}x_{i-1}, \beta_i I)$ \citep{ho-2020}, we can define $q(x_i|x_0, \alpha_i) \sim N(x_i;\sqrt{\alpha_i}x_0, (1-\alpha_i)I)$, where $\alpha_i = \prod\limits_{j=1}^{i}\beta_j$. We denote the perturbed data distribution as $q(\tilde{x};\alpha_i) = \int_\Omega p(x_i,i)q(\tilde{x};x)dx$. A variational Markov chain in the reverse direction is used to sample the denoised image and is given by $q_\theta(x_{i-1}|x_i) \sim N(x_{i-1};\frac{1}{\sqrt{1-\beta_i}}(x_i + \beta_i \tilde{s}_\theta(x_i,i)),\beta_i I)$. The specific loss function to train DDPM is known and will not be repeated \citep{jsong-2020}. The sample of the denoised image is sampled from the reverse Markov chain \citep{ysong-2019}.

\subsection{Fokker-Planck equation for the log-density}
Consider the forward SDE \eqref{eq:forward_sde} and the reverse SDE \eqref{eq:reverse_diffusion}. We observe that the only variable that is unknown in this coupled equation is the score of the distribution $\nabla_x \log p(x(t),t)$. In this paper, we solve the unknown score dynamics using finite difference methods. The numerical solution to the FP equation allows approximate the score of the underlying distribution through time and space. This allows us to sample the corruption in the image $x(t)$ at any given point in time before we train our neural network. We drop the dependence on $x,t$ to simplify notation, when it is clear. We can derive the log-density form of the Fokker-Planck equation by using the relationship $\frac{\partial \log p}{\partial t} = \frac{1}{p}\frac{\partial p}{\partial t}$. Substituting in the original Fokker-Planck equation we get:
\begin{equation}
   \frac{\partial\log p}{\partial t} = -\frac{1}{p}p_t = -\nabla_x F - F\frac{1}{p}\nabla_x p,
\end{equation}
where $\nabla_x$ is the divergence of the distribution with respect to $x$, i.e. $\nabla_x p = \sum_{j=1}^{D}\frac{\partial p}{\partial x_j}$. We can substitute in the relationship $\nabla_x \log p = \frac{1}{p}\nabla_x p$ to get:
\begin{equation}
    \frac{\partial\log p}{\partial t} = -\nabla_xF - F\nabla_x\log p.
\end{equation}
We let $m\equiv \log p$ and substitute $F$ to get the non-linear PDE given by:
\begin{align}
    m_t &= -\nabla_x (f-\frac{1}{2}g^2\nabla_x m) - (f-\frac{1}{2}g^2\nabla_x m) \nabla_x m \nonumber\\ &= -\nabla_x f + \frac{1}{2}g^2\nabla_x^2 m - f\nabla_x m + \frac{1}{2}g^2 (\nabla_x m)^2.
    \label{eq:log-FP}
\end{align}
Equation \eqref{eq:log-FP} is difficult to solve analytically as we have a non-linear first derivative term $(\nabla_x m)^2$. We resolve the nonlinearity by solving the problem numerically. We propose a semi-explicit scheme to resolve the non-linearity. We use linearize the equation by assuming we have an approximation to $\nabla_x m$. Applying the linearization gives us the following PDE:
\begin{equation}
    m_t = -\nabla_x f + \frac{1}{2}g^2\nabla_x^2 m - f\nabla_x m + \frac{1}{2}g^2 \nabla_x m^{-}\nabla_x m,
\end{equation}
where $\nabla_x m^{-}$ is an approximation to $\nabla_x m$. Our approach is free from assumptions on the regularity of the log-density FP equation. Note, when regularity conditions are met, we can derive a PDE of the score \citep{lai-2022}. Our approach solves the log-density FP equation directly and we compute the score directly.

\subsection{Discretizing the Fokker-Planck equation}
To solve the log-density FP equation numerically, first we discretize \eqref{eq:log-FP} using finite difference approximation. Before discretization, we vectorize and normalize the image $x(t)$ such that it has the size $D = HW$. We discretize the continuous variables into discrete domains. Let $n = 1,..., N$ and the time stepsize  $\Delta t = T/N$. Then the discretized points in time is given by $n\Delta t$. We use the standard five-point stencil to discretize the log-density distribution in image space.
        
%
        
        
        




 A standard five-point stencil on a $5\times 5$ grid is used to construct the FDM approximation. Formally, Let $x = 1,..., H$ and $y = 1, ..., W$ be the discretized image with stepsize over the pixels for each channel given by $\Delta x = \Delta y$.  The partial derivatives of the log-density distribution is approximated as follows:
\begin{align}
    m_t &\approx \frac{m_{i,j}^{n} - m_{i,j}^{n-1}}{\Delta t},\nonumber \\
    \nabla_x m &\approx \frac{m_{i+1,j}^{n} - m_{i-1,j}^{n} + m_{i,j+1}^{n} - m_{i,j-1}^{n}}{2},\nonumber \\
    \nabla_x^2 m &\approx m_{i+1,j}^{n} + m_{i,j+1}^{n} - 4m_{i,j}^{n} + m_{i-1,j}^{n} + m_{i,j-1}^{n}.
\end{align}
Similarly, the divergence of $f$ can be approximated as:
\begin{equation}
    \nabla_x f \approx \frac{f_{i+1,j}^{n} - f_{i-1,j}^{n} + f_{i,j+1}^{n} - f_{i,j-1}^{n}}{2}.
\end{equation}
Substituting the approximations into \eqref{eq:reverse_diffusion}, we discretize \eqref{eq:log-FP} as follows:
\begin{align}
    \frac{m_{i,j}^{n} - m_{i,j}^{n-1}}{\Delta t} &= -\frac{f_{i+1,j}^{n} - f_{i-1,j}^{n} + f_{i,j+1}^{n} - f_{i,j-1}^{n}}{2} \nonumber \\ 
    &+ \frac{1}{2}(g^n)^2(m_{i+1,j}^{n} + m_{i,j+1}^{n} - 4m_{i,j}^{n} + m_{i-1,j}^{n} + m_{i,j-1}^{n})\nonumber \\ 
    & \hspace{0.45cm} - (f^n_{i,j} - \frac{1}{2}(g^n)^2\Delta)(\frac{m_{i+1,j}^{n} - m_{i-1,j}^{n} + m_{i,j+1}^{n} - m_{i,j-1}^{n}}{2})^2.
\end{align}

To ensure that we have a good approximation to the score, we use fixed-point iteration to iterate over multiple iterations indexed by $k$ until $m$ converges to its numerical solution. Let $\Delta \tilde{m}^{(k-1),n}_{i,j}$ be the approximation to the score function from the previous iteration $k-1$. Then the non-linear form can be linearized as follows:
\begin{equation}
    (\frac{m_{i+1,j}^{(k),n} - m_{i-1,j}^{(k),n} + m_{i,j+1}^{(k),n} - m_{i,j-1}^{(k),n}}{2})^2 \approx \Delta \tilde{m}^{(k-1),n}_{i,j}(\frac{m_{i+1,j}^{(k),n} + m^{(k),n}_{i,j+1} - m_{i-1,j}^{(k),n}-m_{i,j-1}^{(k),n}}{2}).
\end{equation}
Additionally we denote $\Delta f^{(k),n}_{i/2,j/2} = (f_{i+1,j}^{(k),n} + f_{i,j+1}^{(k),n} - f_{i-1,j}^{(k),n} - f_{i,j-1}^{(k),n})/2$. We construct an implicit finite difference scheme to approximate $m^{(k),n}_{i,j}$ as:
\begin{align}
    \label{eq:log_FP_disc}
    m^{(k),n}_{i,j} &+ \Delta f^{(k),n}_{i/2,j/2} - \frac{1}{2}(g^{(k),n})^2(m_{i+1,j}^{(k),n} + m_{i,j+1}^{(k),n} - 4m_{i,j}^{(k),n} + m_{i-1,j}^{(k),n} + m_{i,j-1}^{(k),n})\\ 
    & \hspace{0.45cm} + \frac{1}{2}(f^{(k),n}_{i,j} - \frac{1}{2}(g^{(k),n})^2\Delta \tilde{m}^{(k-1),n}_{i,j})(m_{i+1,j}^{(k),n} - m_{i-1,j}^{(k),n} + m_{i,j+1}^{(k),n} - m_{i,j-1}^{(k),n}) = m^{(k),n-1}_{i,j}. \nonumber
\end{align}
To solve \eqref{eq:log_FP_disc}, we first form the $HW\times HW$ sparse block coefficient matrix $A$ as:
\begin{equation}
    A = \left[\begin{array}{ c | c | c| c| c}
    D & B & & & \\
    \hline
    C & D & B & & \\
    \hline
     & \ddots & \ddots & \ddots\\
    \hline
     & & C & D & B\\
    \hline
     & & & C & D
  \end{array}\right],
\end{equation}
The coefficient block matrix $A$ is constructed from three different matrices given by:
\begin{equation}
    D = \begin{bmatrix}
        b & a &  &  & \\
        c & b & a &  &  \\
         & \ddots & \ddots & \ddots &  \\
         &  &c & b & a\\
         &  &  & c & b
    \end{bmatrix},
    B = \begin{bmatrix}
        a &  &  &  & \\
         & a &  &  &  \\
         &  & \ddots &  &  \\
         &  & & a & \\
         &  &  &  & a
    \end{bmatrix},
    C = \begin{bmatrix}
        c &  &  &  & \\
         & c &  &  &  \\
         &  & \ddots &  &  \\
         &  & & c & \\
         &  &  &  & c
    \end{bmatrix},
\end{equation}
where $a = (f^{(k),n}_{i,j} - \frac{1}{2}(g^{(k),n})^2\Delta m^{(k-1),n}_{i,j})\Delta t + \frac{1}{2}(g^{(k),n})^2 \Delta t$, $b = 2(g^{(k),n})^2\Delta t$ and $c = (f^{(k),n}_{i,j} - \frac{1}{2}(g^{(k),n})^2\Delta m^{(k-1),n}_{i,j})\Delta t - \frac{1}{2}(g^{(k),n})^2 \Delta t$. All empty entries are zeros. We also construct the vector $B$ of size $HW$ as:
\begin{equation}
    B = \begin{bmatrix}
        m^{(k),n-1}_{i,j}\\
        \vdots\\
        m^{(k),n-1}_{i,j}
    \end{bmatrix}.
\end{equation}
Since $A$ is sparse, we can efficiently solve the linear system $Am=B$ for $m$ where $m$ is a vector of unknown log-densities. Note that our discretization scheme needs to be solved over time.

Solving the system of equations efficiently is non-trivial as standard Gaussian elimination requires full evaluation of the matrix. To achieve efficiency, we take advantage of the sparsity in our approach and use sparse Gaussian elimination to efficiently solve for $m$ at all timesteps \citep{saad-2003}. We use the solution $m$ to compute an approximation of $\nabla_x m$. We compute the score by central differencing this ensures second order accuracy. After $K$ iterations, the approximation to the true $\nabla_x m$ is given by:
\begin{equation}
    \nabla_x m \approx \frac{m^{(K),n}_{i+1,j} + m^{(K),n}_{i,j+1} - m^{(K),n}_{i-1,j} - m^{(K),n}_{i,j-1}}{2}.
\end{equation}
Note, we use zero padding at borders of the image. After computing the numerical approximation to the score. The initial condition is given by the initial density distribution of the image. We use a tree implementation with a linear kernel using scott's algorithm to compute the initial density by kernel density estimation.

We present an algorithm to approximate the score prior to training the score matching network. Algorithm \ref{alg:score_approx_labels} constructs the numerical solution to the FP equation using a finite difference approximation. We use policy-iteration to iterate the solution to the FP equation given the semi-explicit score function as the policy. We allow the algorithm to run until convergence; i.e. errors are less than a set tolerance. This ensures the solution is close to the true solution. Note this is a one time computation performed before the training of the score-matching network.

\begin{algorithm}[htbp!]
\caption{Pre-training: Policy iteration to approximate score from log-density FP equation. We denote the last iteration as $K$.}\label{alg:score_approx_labels}
\begin{algorithmic}
\Require $m^{(k),0}_{i,j},\textnormal{initial guess of }m^{(k-1)}, \nabla_x m^{k-1}, f\textnormal{ at all }\{n, i,j\}, g\textnormal{ at all }\{n\}, tol$
\While{$error > tol$}
\While{$n \in [1, N]$}
\State $\textnormal{Construct }A^{(k),n}, B^{(k),n}$
\State $\textnormal{Solve the system for }m^{(k),n}\textnormal{ with }\nabla_x m^{(k-1),n}$
\EndWhile
\State $\textnormal{Compute and update }\nabla_x m^{(k)}$
\State $\textnormal{Update }error = ||m^{(k)} - m^{(k-1)}||_2$
\EndWhile
\State $\textnormal{Output the approximation }\nabla_x m^{(K)}$
\end{algorithmic}
\end{algorithm}

\subsection{Training the diffusion model}
We use the numerical solution of the log-density FP equation to compute the score and aid the score-matching network in efficiently learning the true score function. To train the score-matching network, first, we embed the score from Algorithm \ref{alg:score_approx_labels} into an image by solving \eqref{eq:probability_flow} forward in time. We discretize \eqref{eq:probability_flow} and sample at timestep $n$. Let
\begin{equation}
    a^{n-1}_{i,j} = (f^{n-1}_{i,j} - \frac{1}{2}(g^{n-1})^2)\nabla_x m^{n-1}_{i,j},
\end{equation}
be the average score between timesteps $n$ and $n-1$. This results in a smoother approximation for sampling. We sample the image at $n$ for each $\{i,j\}$ following the transport model:
\begin{equation}
    x^n_{i,j} = x^{n-1}_{i,j} + a^{n-1}_{i,j}\Delta t.
\end{equation}
In contrast with \citep{boffi-2022}, we do not approximate score at each timestep using a neural network. Instead we assume the numerical solution to the PDE is close to the true score and we embed the score information into the image. This is a form of label embedding in the feature space. By perturbing the score embedded image with $\lambda(t)$, $s_\theta(x,t)$ can efficiently learn the score function. This is because the denoising loss function allows the neural network to map the perturbed image to learn the score of the diffused image.

\begin{figure}[htbp]
    \centering
    \includegraphics[ width = \textwidth]{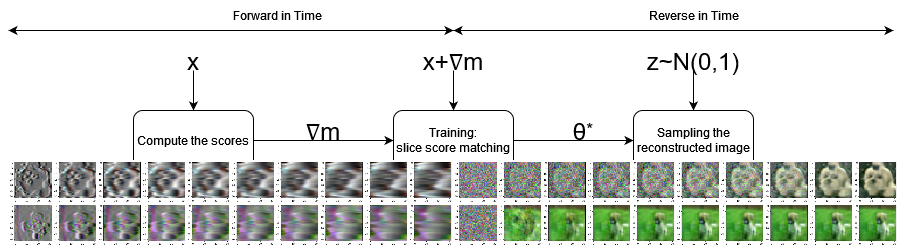}
    \caption{Illustration of training pipeline with score embedding.  }
    \label{fig:score_embed}
\end{figure}
Figure \ref{fig:score_embed} provides a visual summary of our framework. First, we compute the score using finite difference methods. Then, the computed score is passed to the image to aid the neural network to learn the score of the image. An initial noise is given to the sampler and a denoised image is generated efficiently using an ODE solver.

Formally, let $s_\theta(x, t)$ be the score-matching neural network with the score embedded image as an input. To train the score-matching network we use an equivalent denoising loss given by \citep{boffi-2022,vincent-2011}:
\begin{equation}
    \theta^* = \arg\min\limits_{\theta}\{\mathbb{E}_t[\frac{1}{2g(t)^2}\int_{\Omega}||s_{\theta}(x,t)||_2^2 + (g(t))^2 \nabla_x (s_{\theta}(x,t)^\top s_{\theta}(x,t))p(x,t)dx]\},
    \label{eq:denoise_loss_2}
\end{equation}
where $\Omega$ is the domain of the image distribution. Equation \eqref{eq:denoise_loss_2} is difficult to resolve as we need to compute the divergence of a quadratic score term. Instead, let $z \sim N(0,1)$. We approximate the loss \eqref{eq:denoise_loss_2} using sliced score matching as follows \citep{boffi-2022}:
\begin{equation}
    \theta^* = \arg\min\limits_{\theta}\{\mathbb{E}_t[\frac{1}{2g(t)^2}\int_{\Omega}||s_{\theta}(x + \lambda(t)z,t) + z/\lambda(t)||_2^2 p(x,t)dx]\},
\end{equation}
where $\lambda(t)$ is  the perturbation of the image. Intuitively, this allows the score function to learn the score embedded in the image through perturbations. If we assume $\min_{\theta}\{\mathbb{E}_t[\int_{\Omega}||s_{\theta}(x + \lambda(t)z,t) + z/\lambda(t)||_2^2 p(x,t)dx]\} \rightarrow 0$, then we can we multiply $\lambda(t)$ out from the denominator of $z/\lambda(t)$ which improves stability when learning. This gives us the following loss function:
\begin{equation}
    \theta^* = \arg\min\limits_{\theta}\{\mathbb{E}_t[\frac{1}{2g(t)^2}\int_{\Omega}||s_{\theta}(x + \lambda(t)z,t)\lambda(t) + z||_2^2 p(x,t)dx]\},
    \label{eq:final_loss}
\end{equation}

We present the following algorithm to train a score-matching network with label embedding as follows:

\begin{algorithm}[htbp!]
\caption{Training: Label embedding in the feature space}\label{alg:training}
\begin{algorithmic}
\Require $\nabla_x m, x, \lambda, epoch, Optimizer, Model$
\For{$e \in [1, epoch]$}
\For{$n \in [1, N]$}
\State $z \sim N(0,1)$
\State $x^n = x^{n-1} + \alpha^{n/2}\Delta t$
\State $x_{\varepsilon} = x^n + \lambda(n\Delta t)z$
\State $s_{\theta}(x_{\varepsilon},n\Delta t) = Model(x_{\varepsilon},n\Delta t)$
\EndFor
\State $loss \leftarrow \textnormal{Equation \eqref{eq:final_loss}}$
\State $Optimizer(loss)\textnormal{ to backpropagate errors.}$
\EndFor
\end{algorithmic}
\end{algorithm}

Algorithm \ref{alg:training} takes as inputs the score $\nabla_x m$, the normalized image $x$, the perturbation $\lambda$, the number of epochs, the score-matching network model, and the optimizing algorithm as inputs. Over each epoch, we sample the score embedded $x^n$ and perturb it. The perturbed score embedded image is then passed as the input to the score-matching network. Note, mini-batching can easily be implemented into Algorithm \ref{alg:training}. The optimizer we use to train our model is the Adam optimizer \citep{kingma-2014} with a learning rate of $10^{-3}$. To sample from the score we use the ODE sampling for efficient denoising \citep{jsong-2020}.

\section{Experiments}
In this section, we present our numerical results to demonstrate the effectiveness of our method.  We train our neural network on the standard CIFAR 10, ImageNet and the CelebA dataset. Conditional sampling from a dataset refers to the input of the sampler. In this case, the input is an image from the dataset with noise. Unconditional sampling generates the denoised image from the input $z\sim N(0,1)$. We perform two types of experiments. The first experiment is the denoising of one image with unconditional sampling. In this example we demonstrate the improved quality and performance gained by our method for a single image. The second experiment is the denoising performed on multiple images with conditional CIFAR10. In this experiment, we demonstrate the performance of our model when it is trained on multiple images. We also demonstrate our models performance on both conditional ImageNet and unconditional CelebA datasets. We perform our experiments on a single Nvidia Tesla P100 GPU with 16 GBs of memory. The limitation of a single GPU is a bottleneck to our experimentation and limits resolution of images we can use.

The DDPM and DDIM implementation used in the experiments are pulled directly from the github repository of the authors \citep{jsong-2020, ysong-2020}. Both models use a U-net \citep{ronneberger-2015} coupled with an attention network \citep{vaswani-2017}. The U-net is composed of Resnet blocks with temporal embedding layers instead of pooling to allow the network to learn the temporal dynamics. In our model, we use linear time embedding. We found the linearity allows for the time embedding aids in the unconditional generation of images and helps avoid confusion values.

\subsubsection*{Quality comparisons}
To measure quality, we use the mean-squared error (MSE) and structural similarity index metric (SSIM). Both metrics are used widely in image processing literature. In our paper, we also use training time as a secondary metric to demonstrate that our method aids in training in a significant way.  We define them more formally here. Let $y$ be the ground truth and $\tilde{y}$ be the approximation of $y$. Then the MSE is defined as:
\begin{equation}
    MSE(y,\tilde{y}) = \mathbb{E}[\sum_{i}\sum_{j} (y_{i,j} - \tilde{y}_{i,j})^2].
\end{equation}
The SSIM measures the perceived change in the structure, luminence and contrast of an image. Let $L$ denote the dynamic range of the image, then the luminence of an image is measure as \citep{Wang-2004}:
\begin{equation}
    l(y,\tilde{y}) = \frac{2\mu_y\mu_{\tilde{y}} + c_1}{\mu_y^2 + \mu_{\tilde{y}}^2 + c_1},
\end{equation}
where $\mu_y$ is the pixel sample mean of $y$, $\mu_{\tilde{y}}$ is the pixel sample mean of $\tilde{y}$, and $c_1 = 0.01 L$. The contrast is measured as:
\begin{equation}
    c(y,\tilde{y}) = \frac{2\sigma_y\sigma_{\tilde{y}} + c_2}{\sigma_y^2 + \sigma_{\tilde{y}}^2 + c_2},
\end{equation}
where $\sigma_{y}^2$ is the pixel sample variance of $y$, $\sigma_{\tilde{y}}^2$ is the pixel sample variance of $\tilde{y}$, and  $c_2 = 0.03L$. The change in structure is measured by:
\begin{equation}
    s(y,\tilde{y}) = \frac{\sigma_{y\tilde{y}} + c_3}{\sigma_y\sigma_{\tilde{y}} + c_3},
\end{equation}
where $c_3 =c_2/2$, $\sigma_{y\tilde{y}}$ is the covariance between $y$ and $\tilde{y}$. Then the SSIM is given by:
\begin{equation}
    SSIM(y,\tilde{y}) = l(y,\tilde{y})^\alpha c(y,\tilde{y})^\beta s(y,\tilde{y})^\gamma,
\end{equation}
where $\alpha = \beta = \gamma = 1$.

\subsection{Main Results}
In this section, we present the main results of our experiments. We demonstrate the efficiency gains of our proposed method. In the first experiment we look at the performance of our method compared to DDPM and DDIM for a single image case. We found that DDIM gave inconsistent results for multiple images so we omit comparison with DDIM for multiple images. We train each model until they reach a specified SSIM and report their MSE and training time. In the second experiment we make a similar comparison using multiple images. The time results we report is the total training time. For our method, this means we add the time it takes to solve the sparse systems of equations. We train the models to reach a specified SSIM level.
\begin{table}[htbp]
    \centering
    
    \begin{tabular}{|c|c|c|c|c|}
\hline 
 Methods& SSIM&MSE &Training time (s) &Speed up\\ \hline\hline 
     Proposed method &0.99&   0.0009&26.98 & 1\\ \hline 
     DDPM &0.99& 
     0.0006&139.63 &5.17\\ \hline 
      DDIM&0.99& 0.0006&182.53 &6.77\\ \hline\end{tabular}

    \begin{tabular}{|c|c|c|c|c|}
\hline 
 Methods& SSIM&MSE &Training time (s) &Speed up\\ \hline\hline 
     Proposed method &0.98&   0.0015&16.07 &1\\ \hline 
     DDPM &0.98& 
     0.0012&136.91 &8.52\\ \hline 
      DDIM&0.98& 0.0011&182.29 &11.34\\ \hline\end{tabular}
    
    \begin{tabular}{|c|c|c|c|c|}
\hline 
 Methods& SSIM&MSE &Training time (s) &Speed up\\ \hline\hline 
     Proposed method &0.95&   0.0044&9.75 &1\\ \hline 
     DDPM &0.95& 
     0.0033&131.09 &13.44\\ \hline 
      DDIM&0.95& 0.0028&181.57 &18.62\\ \hline\end{tabular}

    \caption{We tabulate the timing results for our proposed method, DDPM and DDIM for three different SSIMs $0.99$ (top), $0.98$ (middle) and $0.95$ (bottom). The speed up column is a ratio of training times between the method and our proposed method.}
    \label{tab:timing_table_cifar1}
\end{table}

\subsection{Single image denoising}
We compare the performance of our method to other methods for the single image denoising task. This experiment is performed on unconditional CIFAR10 for $32\times32$ images and unconditional CelebA for $64\times 64$ images. The number of epochs trained and the learning rate varied for each model as we tuned each model to achieve the maximum SSIM.

\begin{figure}[htbp]
    \centering
    \includegraphics[width=0.8\textwidth]{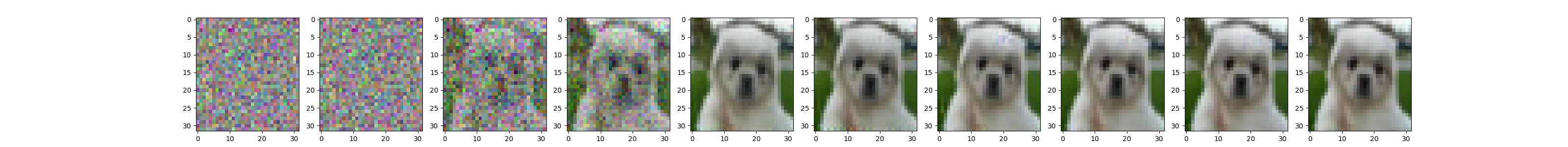}
    \includegraphics[width=0.8\textwidth]{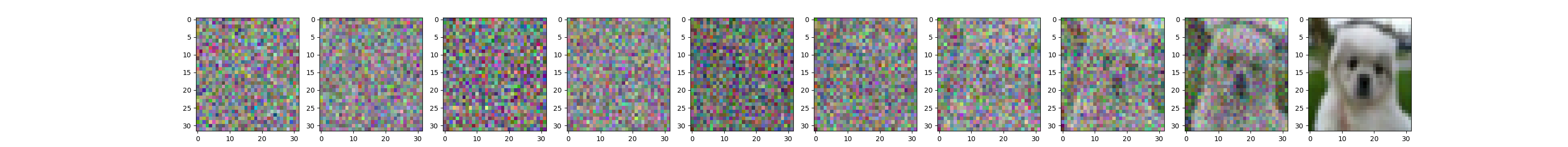}
    \includegraphics[width=0.8\textwidth]{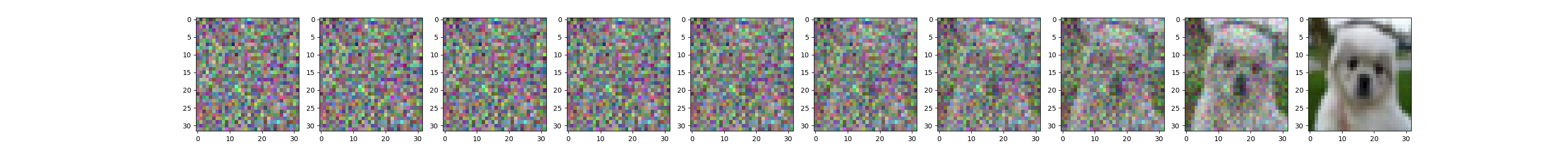}
    \caption{Denoising an image of a puppy sampled from unconditional CIFAR10 using our proposed method (top), DDPM (middle) and DDIM (bottom). We sample $10$ timesteps during the sampling to demonstrate the denoising process.}
    \label{fig:cifar1}
\end{figure}

\begin{figure}
    \centering
    \includegraphics[width=\textwidth]{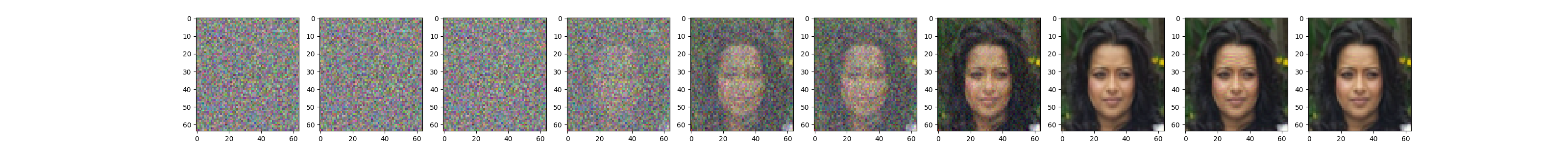}
    \includegraphics[width=\textwidth]{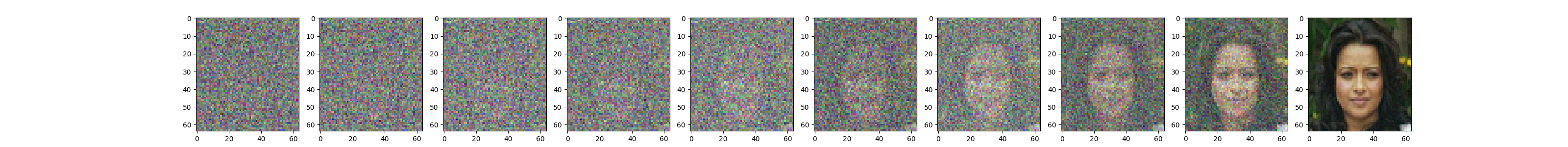}
    \caption{Denoising an image of a female celebrity sampled from unconditional CelebA using our proposed method (top) and DDPM (bottom). We sample $10$ timesteps during the sampling to demonstrate the denoising process.}
    \label{fig:celeb1}
\end{figure}

We summarize the training time required for the neural network to learn to denoise CIFAR10 for SSIM levels of $\{0.99, 0.98, 0.95\}$ in Table \ref{tab:timing_table_cifar1}. In Table \ref{tab:timing_table_cifar1} we observe that good denoising performance occurs for DDPM and DDIM only after it has been sufficiently trained. For our proposed method we see good results early into training. This suggests the neural network learns the features more gradually in a targeted way as it trains. We see improvements of $5.14 - 18.62$ times speed up in training. We can observe the corresponding denoised image in Figure \ref{fig:cifar1}. Figure \ref{fig:cifar1} shows the model gradually denoising the image with acceptable quality.
\begin{table}[]
    \centering

    \begin{tabular}{|c|c|c|c|c|}
\hline 
 Methods& SSIM&MSE &Training time (s) &Speed up\\ \hline\hline 
     Proposed method &0.95&   0.0017&105.41 &1\\ \hline 
     DDPM &0.95& 
     0.0015&601.27 &5.70\\\hline\end{tabular}
    \begin{tabular}{|c|c|c|c|c|}
\hline 
 Methods& SSIM&MSE &Training time (s) &Speed up\\ \hline \hline 
     Proposed method &0.90&   0.0035&71.68 &1\\ \hline 
     DDPM &0.90& 
     0.0027&342.61 &4.78\\\hline\end{tabular}

    \caption{We tabulate the timing results for our proposed method and DDPM for three different SSIMs $0.95$ (top) and $0.90$ (bottom). The speed up column is a ratio of training times between the method and our proposed method.}
    \label{tab:timing_table_celeb1}
\end{table}
We perform similar analysis using the CelebA data set for SSIM levels of $\{0.95, 0.90\}$ in Table \ref{tab:timing_table_celeb1}. We do not make a comparison to DDIM because DDIM would not reach a sufficient SSIM level. All models could not be trained to reach a SSIM above $0.96$. In Table \ref{tab:timing_table_celeb1} we observe improvements of $4.78 - 5.70$ times speed up in training. The gains in CelebA are consistent with the lower bound of CIFAR10. We can observe the corresponding denoised image in Figure \ref{fig:celeb1}. Like Figure \ref{fig:cifar1}, Figure \ref{fig:celeb1} shows the model gradually denoising the image with acceptable quality.

\subsection{Multi-image denoising}
We compare the performance of our method to DDPM in multi image denoising task. This experiment is performed on three images of dogs sampled from conditional CIFAR10 for $32\times32$ images. The number of epochs trained and the learning rate varied for each model as we tried our best to tune each model to maximize the average SSIM. The average SSIM and average MSE are calculated by taking the average of SSIMs and MSEs of the three images.

\begin{figure}
    \centering
    \includegraphics[width=\textwidth]{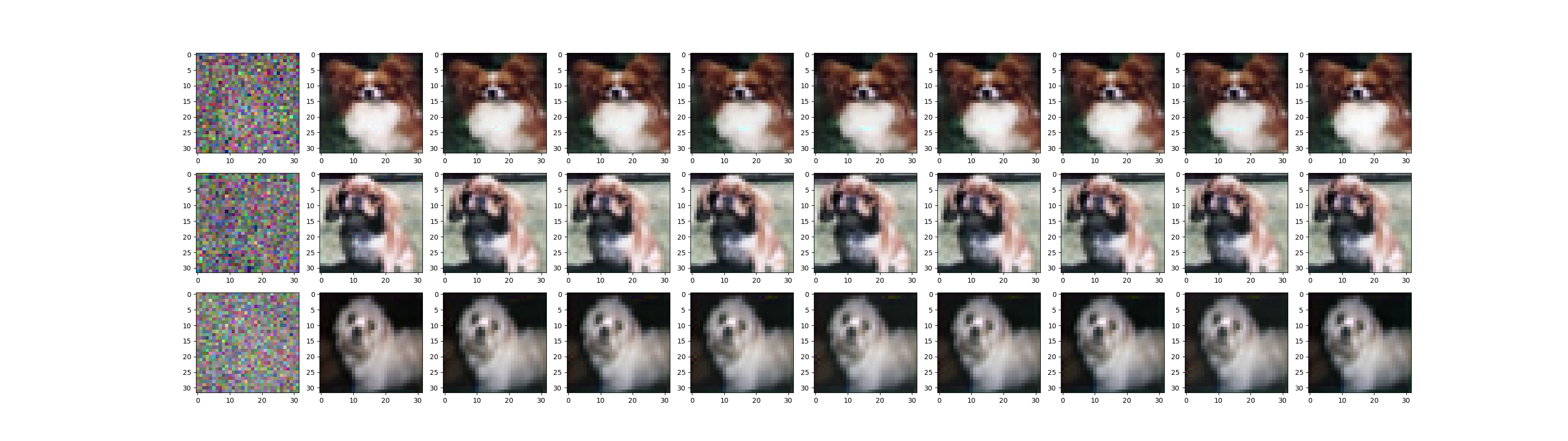}
    \includegraphics[width=\textwidth]{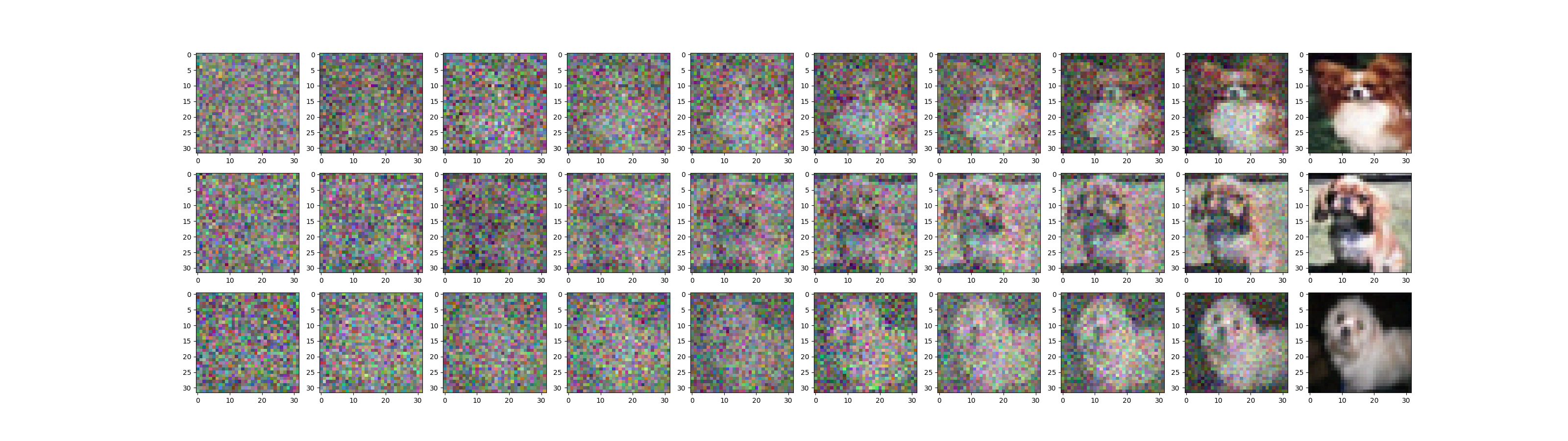}
    \caption{Denoising images of three dogs sampled from conditional CIFAR10 using our proposed method (top) and DDPM (bottom). We sample $10$ timesteps during the sampling to demonstrate the denoising process.}
    \label{fig:cifar3}
\end{figure}

We summarize the training time required for the neural network to learn to denoise CIFAR10 for average SSIM levels of $\{0.95, 0.90\}$ in Table \ref{tab:timing_table_cifar3}. We see improvements of $3.32 - 3.98$ times speed up in training. Figure \ref{fig:cifar3} shows the models gradually denoising the image. However with conditional CIFAR10, our model has already denoised the image after the first few steps. Closer inspection shows colour distortions making the early images look washed out. 
\begin{table}[]
    \centering
    \begin{tabular}{|c|c|c|c|c|}
\hline 
 Methods& avg SSIM&avg MSE&Training time (s) &Speed up\\ \hline\hline
     Proposed method &0.95&   0.0033&46.80 &1\\ \hline 
     DDPM &0.95& 
     0.0037&155.16 &3.32\\\hline\end{tabular}
    
    \begin{tabular}{|c|c|c|c|c|}
\hline 
 Methods& avg SSIM&avg MSE&Training time (s) &Speed up\\ \hline\hline 
     Proposed method &0.90&   0.0078&37.58 &1\\ \hline 
     DDPM &0.90& 
     0.0113&149.72 &3.98\\\hline\end{tabular}
\caption{We tabulate the timing results for our proposed method and DDPM for three different average SSIMs $0.95$ (top) and $0.90$ (bottom). The speed up column is a ratio of training times between the method and our proposed method.}\label{tab:timing_table_cifar3}
\end{table}
Throughout our testing, DDIM produced inconsistent results especially when training on multiple images. Because of these inconsistencies we chose to not compare our results with DDIM for multiple images.

\subsubsection*{Denoising of multiple images from ImageNet}
\begin{figure}
    \centering
    \includegraphics[width=0.75\textwidth]{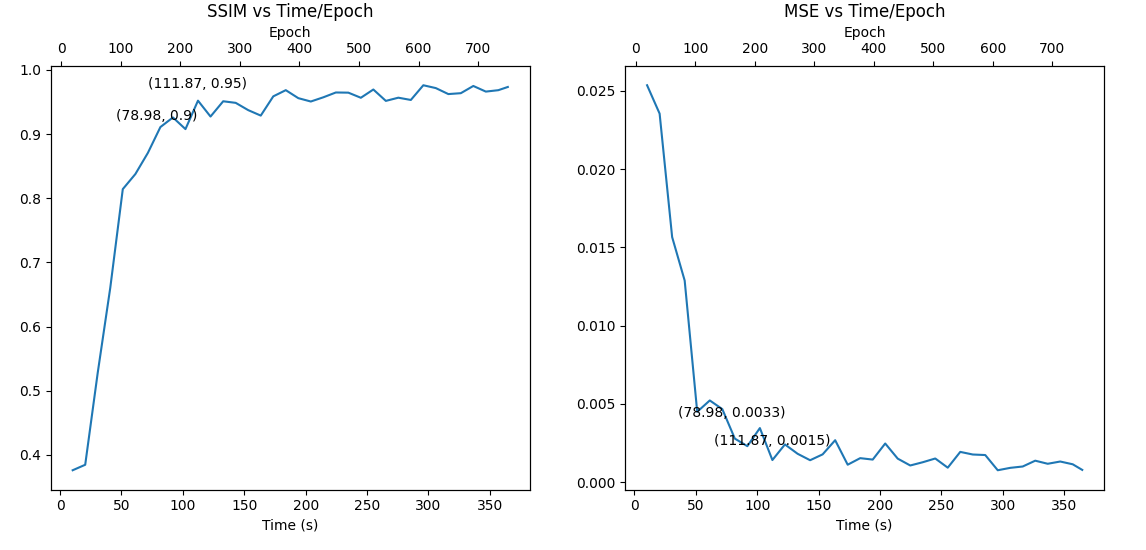}
    \caption{Plot of average SSIM and average MSE curves over training time/epoch.}
    \label{fig:imagenet3_metrics}
\end{figure}

\begin{figure}
    \centering
    \includegraphics[width=\textwidth]{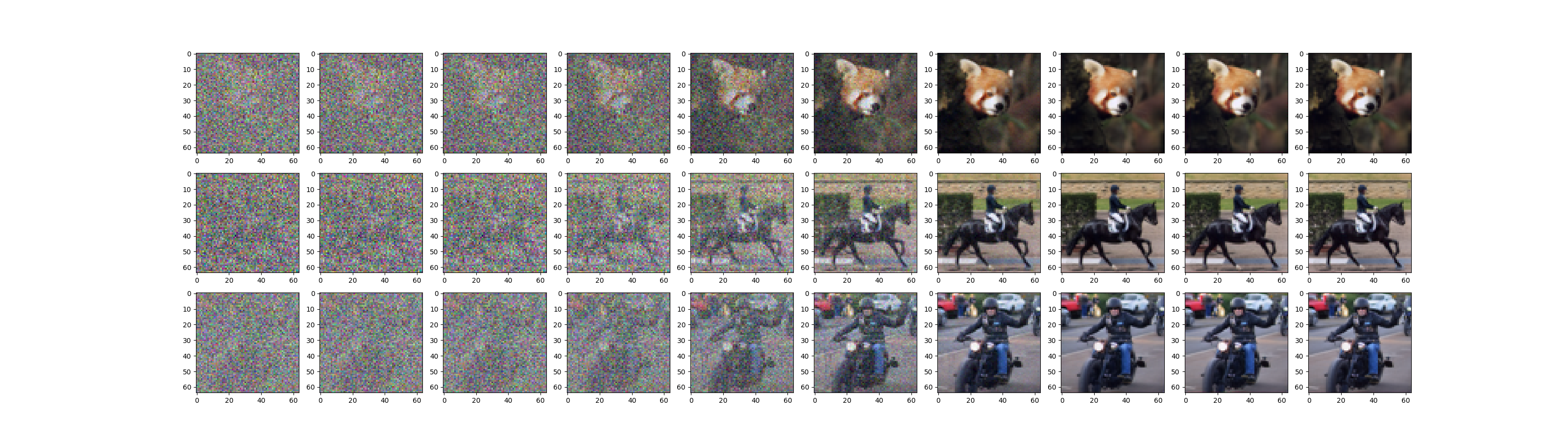}
    \caption{Denoising three images sampled from ImageNet dataset using our proposed method.}
    \label{fig:imagenet3}
\end{figure}

We demonstrate the capability of our model to denoise higher resolution images. We train our model on three images from the ImageNet data set with $64 \times 64$ resolution. Figure \ref{fig:imagenet3_metrics} is a plot of the average SSIM and average MSE achieved over different training times/training epochs. We observe that our method reaches a SSIM of $0.95$ in $111.87$ seconds with an MSE of $0.0015$. We demonstrate the denoised image in Figure \ref{fig:imagenet3}.

\subsection{Progressive generation of multiple images from CelebA}
\begin{figure}[htbp]
    \centering
    \includegraphics[width=\textwidth]{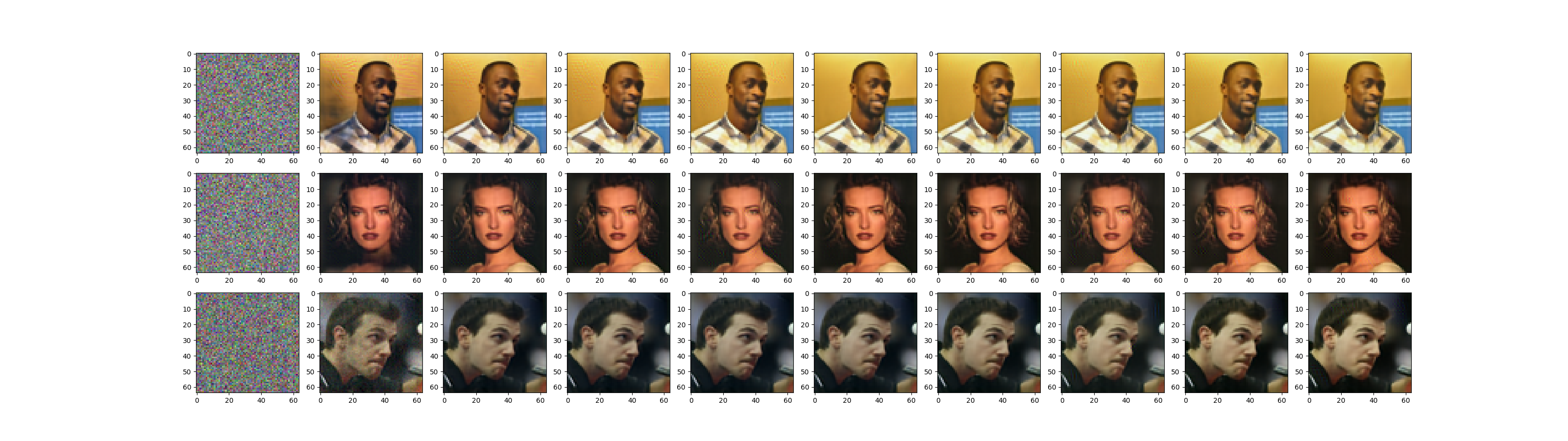}
    \caption{Progressive generation of multiple images sampled from CelebA dataset using our proposed method.}
    \label{fig:celeb3}
\end{figure}
In this section, we demonstrate the generative capabilities of our proposed method. We sample three images from CelebA with a resolution of $64 \times 64$. We use linear time embedding with independent time intervals for each image. This aids our proposed method to be more resilient to the neural network leaking features in between different images resulting in a result similar to conditional sampling. In Figure \ref{fig:celeb3} we show the progressive generation of three celebrities and the scores used for label embedding. As with the single image case we observe that our model is capable of producing recognizable images early but with colour distortions. For this experiment our proposed method reaches a SSIM of $0.95$ in $289.45$ seconds with an MSE of $0.0033$.

\section{Conclusion}
In this paper, we present a method to efficiently denoise an image by improving the training efficiency of score-based diffusion models. We propose to solve the log-density FP equation using sparse numerical methods which is then used to compute the finite difference approximation of the score. This computation step is performed before training. We also propose to embed the pre-computed score into the image using the forward transport equation. This is a form of label embedding into the feature space. Our numerical results show significant improvement in training time for single and multiple images. We demonstrate the effectiveness of our proposed method on CIFAR10, ImageNet and the CelebA dataset. On average, our proposed method achieves a faster training time increase of $3 - 5$ times. This is a marginal gain in efficiency however we expect this to scale well as the image resolution increases.

Future work we will look at extending the approach to videos by solving the score function numerically in higher dimensions. It would also be interesting to develop a framework to learn the dynamic transition between frames and images and the construction of such densities. This would help in making video generation and denoising more efficient.

\section*{Disclosure Statement}
No potential conflict of interest is reported by the author(s).

\section*{Funding}
This work was supported by the Natural Sciences and Engineering Research Council of Canada.

\section*{ORCID}
\begin{description}
\item \textit{Andrew Na}: https://orcid.org/
0000-0002-6162-8171
\item \textit{Justin W.L. Wan}: https://orcid.org/
0000-0001-8367-6337
\end{description}

\bibliographystyle{plainnat}
\bibliography{References}

\end{document}